\documentclass{article}

\usepackage[final]{neurips_2025}

\usepackage[utf8]{inputenc} 
\usepackage[T1]{fontenc}    
\usepackage{hyperref}       
\usepackage{url}            
\usepackage{booktabs}       
\usepackage{amsfonts}       
\usepackage{xspace}

\usepackage{amsmath}        
\usepackage{nicefrac}       
\usepackage{microtype}      
\usepackage{xcolor}         
\usepackage{graphicx}
\usepackage{algorithmic}
\usepackage[linesnumbered, vlined, ruled]{algorithm2e}
\newcommand{\RNum}[1]{\uppercase\expandafter{\romannumeral #1\relax}}
\usepackage{amssymb} 
\usepackage{geometry} 
\usepackage{bm} 
\usepackage{enumitem} 
\usepackage{subfig}

\usepackage{caption}

\title{Co-Win: Joint Object Detection and Instance Segmentation in LiDAR Point Clouds via Collaborative Window Processing}

\author{%
  Haichuan Li \\
  Turku Intelligent Embedded and Robotics Systems lab, Faculty of Technology\\
  University of Turku\\
  Turku, Finland \\
  \texttt{haicli@utu.fi} \\
  \And
  Tomi Westerlund \\
  Turku Intelligent Embedded and Robotics Systems lab, Faculty of Technology \\
  University of Turku\\
  Turku, Finland \\
  \texttt{tovewe@utu.fi} \\
}

\begin{document}

\maketitle

\begin{figure}[h]
    \centering
    \captionsetup{type=figure}
    \includegraphics[width=1\textwidth]{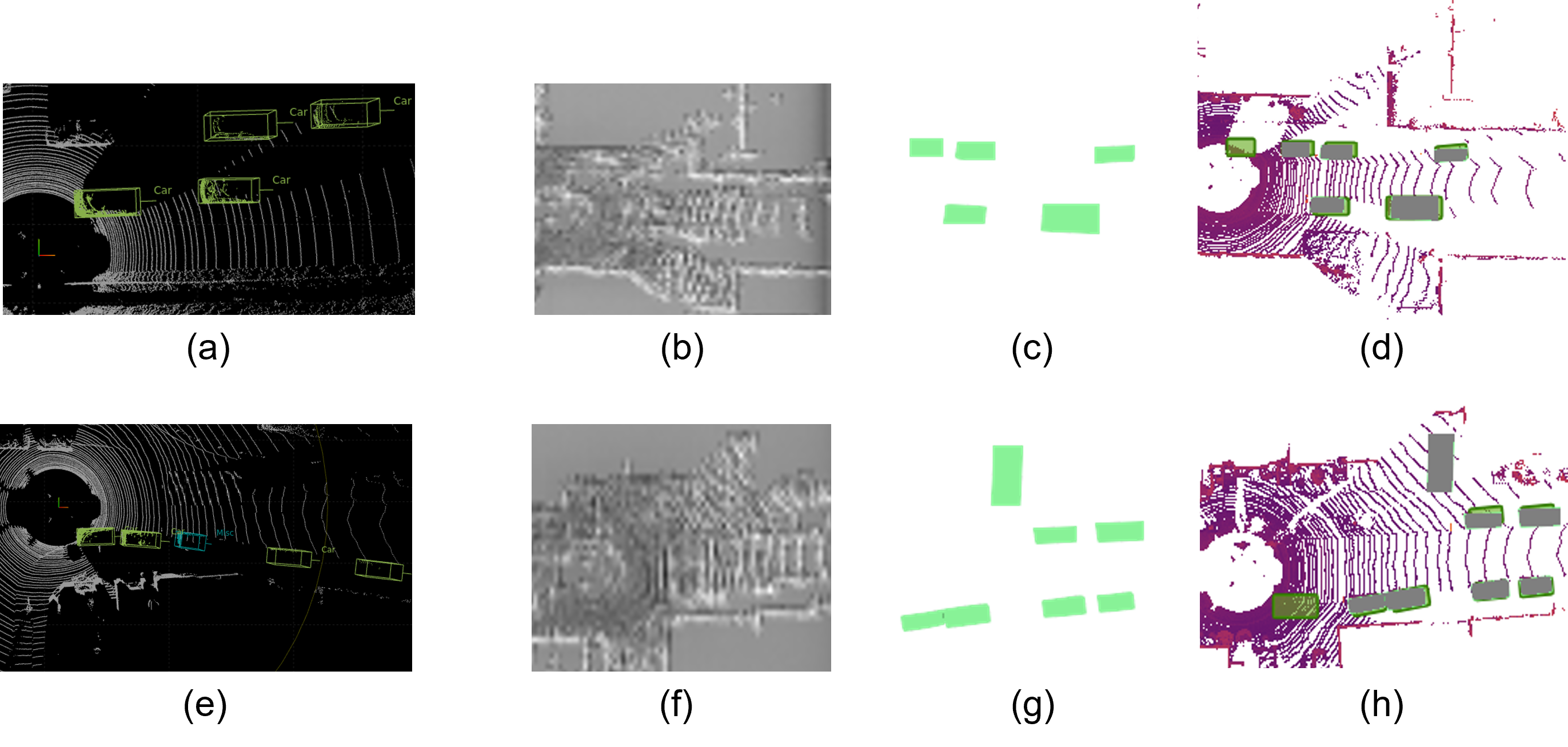}
    \caption{The data transformation process of the Co-Win algorithm is illustrated. Subfigures (a) and (e) are the input to the AFN, representing raw 3D point clouds. Subfigures (b) and (f) depict the bird's-eye view representation after point cloud projection and feature extraction using the SPCN architecture. Subfigures (c) and (g) present the ground truth instance footprint masks. Subfigures (d) and (h) display the output of the mask-based decoder, showing predicted instance masks in bird's-eye view.}
    \label{fig:workflow}
\end{figure}

\begin{abstract}

Accurate perception and scene understanding in complex urban environments is a critical challenge for ensuring safe and efficient autonomous navigation. In this paper, we present Co-Win, a novel bird's eye view (BEV) perception framework that integrates point cloud encoding with efficient parallel window-based feature extraction to address the multi-modality inherent in environmental understanding. Our method employs a hierarchical architecture comprising a specialized encoder, a window-based backbone, and a query-based decoder head to effectively capture diverse spatial features and object relationships.
Unlike prior approaches that treat perception as a simple regression task, our framework incorporates a variational approach with mask-based instance segmentation, enabling fine-grained scene decomposition and understanding. The Co-Win architecture processes point cloud data through progressive feature extraction stages, ensuring that predicted masks are both data-consistent and contextually relevant. Furthermore, our method produces interpretable and diverse instance predictions, enabling enhanced downstream decision-making and planning in autonomous driving systems.
\end{abstract}
	
\section{Introduction}
\label{sec:intro}

Bird's-eye view (BEV) is an innovative strategy in ground vehicle perception systems. BEV offers a comprehensive top-down view of the surrounding environment, facilitating object detection from various perspectives. This detailed positional and semantic information promotes accurate obstacle avoidance and motion planning.
Various sensors, including cameras, LiDAR, and radar, are used to generate BEV representations. LiDAR sensors are widely used in perception tasks due to their ability to obtain precise and reliable 3D information. However, the unstructured nature of LiDAR point clouds poses computational challenges for object detection. The BEV method addresses this challenge by predicting class labels for each pixel in a raster format, thereby avoiding the complexity of generating precise vector contours.
Occlusion, a prevalent issue in BEV object detection, frequently results in partially obscured objects, limiting visibility to the LiDAR-facing surfaces~\cite{10587405}. Consequently, effective BEV generation and 3D object detection must explicitly address this inherent challenge of LiDAR data to achieve high accuracy. 

The common approach to BEV generation involves point cloud preprocessing, feature extraction, and BEV decoding.
Preprocessing strategies for point clouds largely fall into two categories: 3D voxel-based and 2D pillar-based methods~\cite{Zhou2017VoxelNetEL,8954311,Liu2023PointRCNNTM}. While voxel-based methods offer higher accuracy, they suffer from high memory demands. Conversely, pillar-based methods provide greater efficiency at the expense of reduced spatial information capture.
Following preprocessing, a feature extraction procedure is applied to create a feature representation often referred to as a "pseudo-image"~\cite{Wang2018PseudoLiDARFV,Shao2024SparsePD}. This pseudo-image is then utilized in downstream tasks such as object detection, segmentation, and classification.
Finally, a decoder generates the BEV representation from the processed information, serving as the network's output. This decoder predicts the location, shape, and class of objects~\cite{Zong2023TemporalET,Zhao2024MaskBEVTA}.

In this paper, we propose the \textit{Co-Win} algorithm, which utilizes a mask-based approach for object detection and instance segmentation, detecting objects through their footprint masks rather than relying on traditional bounding box regression. Our algorithm performs joint object detection and footprint completion in a single pass, providing precise shape information that conventional bounding box methods cannot capture.
We evaluate the proposed \textit{Co-Win} algorithm on several challenging benchmarks, including KITTI~\cite{Geiger2013}, Waymo Open Dataset~\cite{Sun2020}, and SemanticKITTI~\cite{Behley2019}.
Our \textit{Co-Win} algorithm achieves significant performance gains over existing state-of-the-art methods.
The key contributions of this work are as follows:
\begin{itemize}
    \item An efficient and accurate pre-processing network as encoder for compacting point clouds and extracting geographic information for each LiDAR point cloud cluster.
    \item A parallel processing network (SPCN) incorporating a linear attention mechanism for efficient feature extraction from the pre-processed data.
    \item A novel mask-based decoder architecture for high-fidelity BEV map generation and precise instance segmentation instead of traditional bounding box regression.
\end{itemize}

\section{Related Work}
\label{sec:related_work}

Bird's-eye view (BEV) representation has emerged as a dominant paradigm for 3D perception in autonomous driving scenarios. Early approaches like MV3D~\cite{chen2017multi} and AVOD~\cite{ku2018joint} projected 3D point clouds onto a 2D plane to create a dense representation for object detection. Recent works have focused on generating more informative BEV representations. BEVFusion~\cite{liu2023bevfusion} achieved this through multi-modal fusion, while methods like BEVDet~\cite{huang2021bevdet} and BEVFormer~\cite{li2022bevformer} leveraged image-based features to enhance BEV quality. Unlike these approaches, our Co-Win algorithm generates high-fidelity BEV representations through a dedicated mask-based approach that preserves fine-grained object details.
Feature extraction is a crucial step in BEV generation, addressing the challenges of sparse point cloud data. Early voxel-based methods like VoxelNet~\cite{Zhou2017VoxelNetEL} paved the way for point cloud processing by dividing 3D space into regular voxels. PointPillars~\cite{8954311} introduced efficient pillar feature encoding, while PointRCNN++~\cite{Liu2023PointRCNNTM} enhanced accuracy through distance bin-based encoding. Dynamic voxelization~\cite{Zhou2019EndtoEndMF} deploys initial sparse voxel and pillar feature generation. More recent approaches like FSD~\cite{10506794} employ sparse voxel feature extractors to generate feature maps, while PiSFANet~\cite{10595495} introduces a real-time and scale-aware network with a robust pillar feature extraction encoder. While voxel-based methods offer higher accuracy, they often suffer from high memory demands; conversely, pillar-based methods provide greater computational efficiency but capture less spatial information.
Point cloud object detection methods can be broadly categorized into three types: point-based, voxel-based, and BEV-based approaches. Point-based methods like PointRCNN~\cite{shi2019pointrcnn} and 3DSSD~\cite{yang20203dssd} operate directly on raw points, preserving geometric details but suffering from computational inefficiency with large point clouds. Voxel-based methods such as VoxelNet~\cite{Zhou2017VoxelNetEL} and SECOND~\cite{yan2018second} discretize 3D space into regular grids for convolutional processing, trading precision for efficiency. BEV-based methods like PointPillars~\cite{8954311} and CenterPoint~\cite{yin2021center} project points onto a 2D plane for efficient processing. Our Co-Win algorithm advances BEV-based detection by combining the efficiency of 2D processing with the precision of mask-based instance representation.
Traditional segmentation approaches classify individual pixels or points, but recent mask-based methods reframe segmentation as predicting coherent groups of pixels/points with associated class labels. Mask2Former~\cite{cheng2021mask2former} pioneered this reframing for 2D images, while methods like MaskPLS~\cite{Marcuzzi2023MaskBasedPL} extended this concept to 3D point clouds by predicting semantic classes and clustering "thing" points for instance segmentation. MPVSS~\cite{Weng2023MaskPF} employs a query-based image segmentor to generate accurate binary masks and class predictions on sparse key frames. In the mapping domain, Mask2Map~\cite{Choi_2024} focuses on predicting classes and ordered point sets of map instances within a scene represented in BEV.
Joint detection and segmentation approaches aim to unify these traditionally separate tasks to improve overall performance through shared feature learning. In the 2D domain, methods like Mask R-CNN~\cite{he2017mask} pioneered this approach by extending Faster R-CNN with a mask prediction branch. For 3D point clouds, PanopticFusion~\cite{narita2019panopticfusion} and DS-Net~\cite{hong2021lidar} demonstrated the benefits of joint detection and segmentation for scene understanding. More recently, BEV-based joint methods like PolarStream~\cite{chen2021polarstream} have shown promising results in dynamic environments. The Co-Win algorithm advances this line of research by performing joint object detection and footprint completion in a single pass, leveraging a unified architecture that optimizes both tasks simultaneously through mask-based representation.

\section{Method}
\label{sec:method}

This section presents the proposed Co-Win algorithm which consists of three main components: an Axis-Fusion Network (AFN) for preprocessing 3D point clouds, a Sub-window Parallel Computing Network (SPCN) for feature extraction, and a mask-based decoder for instance segmentation. Figure~\ref{fig:workflow} illustrates the overall architecture of our pipeline.

\subsection{Axis-Fusion Network (AFN)}
\label{sec:afn}

The cornerstone of our perception system is the Axis-Fusion Network (AFN), a novel encoder architecture designed to transform unstructured point clouds into comprehensive representations while preserving critical geometric information(Figure.~\ref{fig:cowin-encoder}). Unlike conventional approaches that simply voxelize 3D data, our method employs a multi-perspective fusion strategy that maintains directional information.
Given an input point cloud $P = \{p_i | p_i \in \mathbb{R}^4\}_{i=1}^N$, where each point $p_i = [x, y, z, r]$ contains 3D coordinates and reflection intensity, we first filter points within a predefined region of interest:
\[ P_{roi} = \{ p_i \in P \mid x_{min} < p_i^x < x_{max}, y_{min} < p_i^y < y_{max}, z_{min} < p_i^z < z_{max} \} \]
The filtered points are organized into a sparse voxel representation using voxel size $v_x \times v_y \times v_z$. For each non-empty voxel, we compute statistical features including point distribution, mean coordinates, and local density patterns.

\paragraph{Multi-Perspective Feature Extraction}
We introduce a novel perspective-based feature extraction approach that explicitly addresses the inherent limitations of single-view representations by analyzing the point cloud from three orthogonal planes:
\[ F^{XZ} = \text{Conv}_{\text{XZ}}(P_{roi}) \cdot (1 + \text{mean}(\sin(\alpha z), \cos(\alpha x)))\] 
\[F^{YZ} = \text{Conv}_{\text{YZ}}(P_{roi}) \cdot (1 + \text{mean}(\sin(\alpha z), \cos(\alpha y))), \qquad F^{XY} = \text{Conv}_{\text{XY}}(P_{roi}) \]
where $\alpha$ is a learnable angular scaling factor. This trigonometric modulation enhances directional awareness and spatial relationships between points in each projection plane.

\paragraph{Multi-Scale Denoising}
To enhance feature quality, especially for sparse regions and distant objects, we employ a specialized denoising network with multi-scale dilated convolutions:
\[ X_{\text{branch1}} = \mathcal{F}_{\text{dil=1}}(X), \qquad X_{\text{branch2}} = \mathcal{F}_{\text{dil=2}}(X), \qquad X_{\text{branch3}} = \mathcal{F}_{\text{dil=4}}(X)\]
\[X_{\text{denoised}} = \mathcal{F}_{\text{fusion}}(X_{\text{branch1}}, X_{\text{branch2}}, X_{\text{branch3}}) + X \]
This architecture preserves fine-grained structures while effectively suppressing noise through complementary receptive fields.

\paragraph{Geometric Multi-Axis Fusion Mechanism}
The features from different projection planes contain complementary information that must be integrated efficiently. Instead of computationally expensive attention mechanisms, we propose a Geometric Axis Fusion (GAF) approach that leverages physical constraints and statistical methods:
\[ \text{GAF}(F^{XZ}, F^{YZ}, F^{XY}) = \mathcal{S}\left( w_{XZ} \cdot F^{XZ} + w_{YZ} \cdot F^{YZ} + w_{XY} \cdot F^{XY} \right) \cdot \mathcal{C} \]
where $w_i$ are confidence weights determined by feature variance statistics, $\mathcal{S}$ is a statistical fusion function, and $\mathcal{C}$ is a geometric consistency factor derived from inter-feature correlations. This formulation provides computational efficiency while maintaining geometric validity.
The confidence weights are computed as:
$ w_i = \frac{e^{-\text{Var}(F^i)} \cdot \alpha_i}{\sum_j e^{-\text{Var}(F^j)} \cdot \alpha_j} $
where $\alpha_i$ are learnable importance parameters and $\text{Var}(F^i)$ quantifies the feature variance, with lower variance indicating higher confidence.

The geometric consistency factor ensures that the fused features respect physical constraints:
\[ \mathcal{C} = \frac{1}{3}(\text{sim}(F^{XZ}, F^{YZ}) + \text{sim}(F^{XZ}, F^{XY}) + \text{sim}(F^{YZ}, F^{XY})) \]
where $\text{sim}(\cdot,\cdot)$ measures feature similarity, enforcing consistency across different perspectives.

\paragraph{Advanced Height Encoding}

Height information is crucial for distinguishing objects in the 3D scene. We implement a multi-frequency positional encoding inspired by NeRF:
\[ \gamma(z) = \{ \sin(2^l \pi z), \cos(2^l \pi z) \}_{l=0}^{L-1},\qquad \gamma_{\text{weighted}}(z) = \gamma(z) \cdot \exp(-0.5 \cdot [0, 1, \dots, 1]) \]
where $L=6$ frequency bands capture both fine and coarse height variations. This encoding is weighted by frequency importance.

\paragraph{Global Geographic Information Tokens}

In parallel with local feature extraction, we compute Global Geographic Information Tokens (GGIT) that capture scene-level context:
$ \text{GGIT} = \mathcal{T}(F_{\text{global}}, F_{\text{cluster}}, F_{\text{structure}}) $
where $\mathcal{T}$ is a token generator, and the input features encode:
\RNum{1}: $F_{\text{global}}$: Statistical distributions across the entire point cloud.
\RNum{2}: $F_{\text{cluster}}$: Density variations and potential object locations.
\RNum{3}: $F_{\text{structure}}$: Scene topology including estimated ground plane parameters

Each token element is computed through a combination of learned embeddings and context-specific updates:
$ \text{GGIT}_i = E_i + \Delta_i(\mathcal{T}) $
where $E_i$ are learnable token embeddings and $\Delta_i$ are context-dependent updates.

\paragraph{BEV Representation}

The final output combines the fused features into a bird's-eye view representation:
\[ F_{\text{BEV}} = \text{LayerNorm}\left(\mathcal{P}(F_{\text{fused}})\right) \in \mathbb{R}^{H \times W \times C} \]
where $\mathcal{P}$ is a projection function that aggregates height information into channel dimensions while preserving spatial layout in the $x$-$y$ plane.

Through this carefully designed architecture, the AFN preserves crucial geometric information while efficiently transforming raw point clouds into structured representations suitable for downstream perception tasks.

\subsection{Sub-window Parallel Computing Network (SPCN)}
\label{sec:spcn}

The Sub-window Parallel Computing Network (SPCN) serves as the feature extraction backbone of our architecture, receiving the BEV representations produced by the AFN. The SPCN employs a novel sub-window parallel processing approach that significantly reduces computational complexity while maintaining representational power.

\paragraph{Sub-window Partitioning Paradigm}

Unlike conventional transformer architectures that process the entire feature map uniformly, our SPCN decomposes the BEV feature map into non-overlapping sub-windows, enabling localized processing:
\[ \{ S_1, S_2, \ldots, S_K \} = \text{Partition}(F_{\text{BEV}}, M) \]
where $F_{\text{BEV}} \in \mathbb{R}^{H \times W \times C}$ is the BEV feature map, $M$ is the sub-window size, and $K = \frac{H \times W}{M^2}$ is the number of sub-windows. This partitioning strategy allows for efficient parallel computation and reduces memory requirements by a factor of $\mathcal{O}(M^2)$ compared to global attention methods.

\paragraph{Parallel Sub-window Processing}

Each sub-window $S_i$ is processed independently using a specialized Sub-window Block (SWB) structure:
\[ \hat{S}_i = \text{SWB}(S_i, \theta), \qquad \forall i \in \{1, 2, \ldots, K\} \]
where $\theta$ represents the learnable parameters of the block. The SWB consists of multi-head self-attention followed by a feed-forward network:
\[ \text{SA}(Q,K,V) = \text{Softmax}\left(\frac{QK^T}{\sqrt{d}}\right)V, \qquad X' = X + \text{MSA}(\text{LN}(X)), \qquad \hat{X} = X' + \text{FFN}(\text{LN}(X')) \]
where MSA denotes multi-head self-attention, LN is layer normalization, and FFN is a feed-forward network. The parallel nature of this computation allows us to efficiently leverage modern GPU architectures.

\paragraph{Linear Attention for Efficient Processing}

To optimize computational efficiency, our SPCN backbone implements linear attention as the primary attention mechanism. Linear attention reduces complexity from $O(N^2)$ to $O(N)$ for sequence length $N$, making it particularly suitable for resource-constrained applications. While standard attention involves a computationally expensive dot product between queries and keys. Our linear attention formulation leverages the kernel trick:
\[ \text{NormalAttention}(Q, K, V) =\text{Softmax}\left(\frac{QK^T}{\sqrt{d}}\right)V \quad \text{LinearAttention}(Q, K, V) = \phi(Q)\left(\phi(K)^T V\right) \]
where $\phi(\cdot)$ is a kernel feature map that projects inputs into a space where dot products correspond to desired similarity measures. Our implementation employs the ELU+1 feature map:
$ \phi(x) = \text{ELU}(x) + 1 $
This formulation ensures positive values and enables reordering matrix multiplications to achieve linear complexity. By computing $(\phi(K)^T V)$ first, followed by multiplication with $\phi(Q)$, we reduce the computational complexity from $O(N^2d)$ to $O(Nd^2)$, where $d$ is the feature dimension. This optimization enables our SPCN to efficiently process large feature maps without sacrificing representation power, making it particularly suitable for resource-constrained edge devices and real-time perception systems.

\paragraph{Block Sequence and Feature Hierarchy}

The SPCN backbone consists of $L$ sequential layers of Sub-window Block Sequences, each containing multiple SWBs:
$
{B}_l = \{B_{l,1}, B_{l,2}, \ldots, B_{l,D_l}\}
$
where $D_l$ is the depth of the $l$-th layer. Each layer progressively transforms the feature representation, with key properties:

\begin{itemize}
    \item \textbf{Progressive Resolution Reduction:} Each layer reduces spatial resolution by a factor of 2 using patch merging operations: $H_l = \frac{H_{l-1}}{2}, W_l = \frac{W_{l-1}}{2}$
    \item \textbf{Channel Expansion:} As spatial resolution decreases, feature channel dimension increases: $C_l = 2 \cdot C_{l-1}$
    \item \textbf{Dynamic Sub-window Size:} To maintain computational balance, sub-window size adjusts across layers: $M_l = \frac{M_{l-1}}{2}$ for $l > 1$
\end{itemize}

This hierarchical structure enables the network to capture multi-scale contextual information while maintaining computational efficiency.

\paragraph{Feature Stage Generation}

The SPCN produces a multi-scale feature hierarchy $\{F_1, F_2, F_3, F_4\}$ across four stages with progressively decreasing spatial resolution and increasing channel dimension:
$
F_l \in \mathbb{R}^{\frac{H}{2^l} \times \frac{W}{2^l} \times C \cdot 2^l}
$
Features at each scale capture different aspects of the scene:
\RNum{1}: $F_1$ (high-resolution): Fine-grained spatial details.
\RNum{2}: $F_2$ (mid-resolution): Object-part relationships.
\RNum{3}: $F_3$ (mid-low resolution): Object-level semantics.
\RNum{4}: $F_4$ (low-resolution): Global scene context.

\paragraph{Global Geographic Information Integration}

The GGIT tokens from the AFN are integrated into the SPCN through a novel token-to-feature interaction module:
\[
\hat{S}_i = S_i + \text{Proj}(\text{Attn}(S_i, \text{GGIT}))
\]
where Attn represents cross-attention between sub-window features and GGIT tokens, and Proj is a projection function that aligns the dimensions. This integration ensures global contextual awareness within local sub-window processing, effectively combining local geometric precision with global scene understanding.

\paragraph{Computational Complexity Analysis}

The computational advantage of our SPCN comes from the localized attention computation:
\[
\Omega(\text{SPCN}) = K \cdot \mathcal{O}(M^4) = \mathcal{O}(HW \cdot M^2)
\]
compared to the quadratic complexity of global attention:
\[
\Omega(\text{GlobalAttn}) = \mathcal{O}((HW)^2)
\]
For typical values ($H=W=200, M=10$), this represents a 400× reduction in computational complexity, enabling real-time processing on standard hardware while maintaining high representational capacity.


\subsection{Mask-Based Decoder Architecture}
\label{sec:mask-decoder}

The final component of our Co-Win architecture is a mask-based decoder that transforms multi-scale features from the SPCN backbone into instance-level object masks(Fig.~\ref{fig:cowin-decoder}). Unlike traditional bounding box detectors, our decoder directly generates precise object masks, enabling more accurate shape delineation particularly for irregular objects.

\begin{figure}[htbp]
  \centering
  \includegraphics[width=\textwidth]{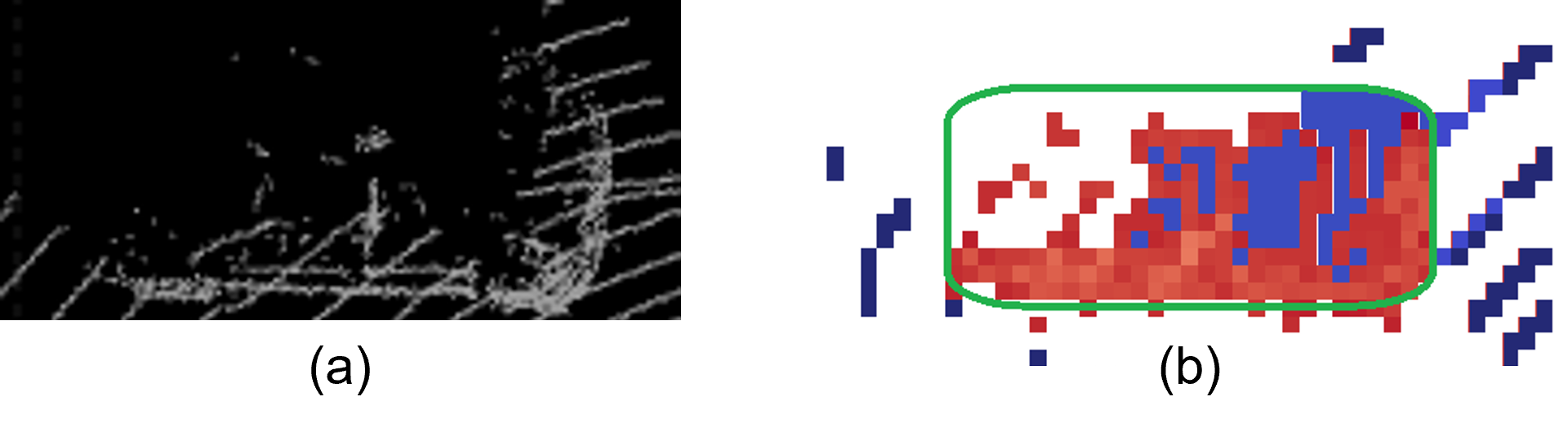}
  \caption{An example of object mask in BEV. The subfigure (a) shows the ground truth and point clouds cluster of a vehicle in the raw 3D point clouds, and subfigure (b) the mask of the vehicle in the BEV.}
  \label{fig:MaskExample}
\end{figure}

\paragraph{Boundary Completion}
To assess object coverage within individual scans, we calculate the ratio of the observed instance mask area (illustrated in heatmap in Fig.~\ref{fig:MaskExample}(b)) to the complete instance mask area (outlined in green). This ratio is computed for all instances across all scans, retaining only the maximum value per instance, representing the scan with the greatest visibility for that instance. This "best-case" approach quantifies the extent to which a single scan can capture the true object boundary under ideal conditions. Instances with limited visibility across all scans (below a predefined threshold) are excluded from this analysis, as their inherently small observed areas would not accurately reflect the method's performance.

\paragraph{Multi-scale Deformable Attention Pixel Decoder}

The decoder first employs a multi-scale deformable attention (MSDA) pixel decoder to progressively integrate features from different levels:
\[
\hat{F}_l = \text{MSDeformAttn}(F_l, \{p_i\}_{i=1}^L, \{F_i\}_{i=1}^L)
\]
where $F_l \in \mathbb{R}^{H_l \times W_l \times C_l}$ is the feature map from level $l$, $\{p_i\}_{i=1}^L$ are learnable sampling offsets, and $\{F_i\}_{i=1}^L$ is the set of all multi-scale features. The deformable attention mechanism adaptively samples points from different spatial locations across all feature levels, enabling the integration of information from various receptive fields.

The sampling operation is mathematically defined as:
\[
\text{MSDeformAttn}(q, p, x) = \sum_{m=1}^M \sum_{k=1}^K A_{mk} \cdot W_m \cdot x_k(\phi_k(p_{mk}))
\]
where $q$ is the query feature, $p$ is the reference point, $A_{mk}$ are attention weights, $W_m$ are projection matrices, $x_k$ represents the feature map of level $k$, and $\phi_k(p_{mk})$ denotes the sampling location in level $k$ for attention head $m$.

\paragraph{Transformer Decoder with Object Queries}

Our transformer decoder processes a set of $N$ learnable object queries $\{q_i\}_{i=1}^N$ that represent potential objects in the scene. Each decoder layer $l$ consists of three primary operations:
\[
\tilde{q}_i^l = q_i^l + \text{MSA}(\text{LN}(q_i^l), \text{LN}(Q^l)) \quad
\hat{q}_i^l = \tilde{q}_i^l + \text{MCA}(\text{LN}(\tilde{q}_i^l), \text{LN}(\hat{F})) \quad
q_i^{l+1} = \hat{q}_i^l + \text{FFN}(\text{LN}(\hat{q}_i^l))
\]

where $\text{MSA}$ is multi-head self-attention, $\text{MCA}$ is multi-head cross-attention, $\text{LN}$ is layer normalization, and $\text{FFN}$ is a feed-forward network. $Q^l = \{q_1^l, q_2^l, \ldots, q_N^l\}$ represents the set of all object queries at layer $l$.

The self-attention mechanism allows object queries to exchange information, capturing relationships between potential objects:
$
\text{MSA}(Q) = \text{Concat}(\text{head}_1, \ldots, \text{head}_h)W^O
$
where each attention head is computed as:
\[
\text{head}_i = \text{Softmax}\left(\frac{QW_i^Q \cdot (QW_i^K)^T}{\sqrt{d_k}}\right) \cdot QW_i^V
\]
The cross-attention enables object queries to gather relevant information from the pixel features:
\[
\text{MCA}(Q, \hat{F}) = \text{Concat}(\text{head}_1, \ldots, \text{head}_h)W^O
\]
with each head calculated as:
\[
\text{head}_i = \text{Softmax}\left(\frac{QW_i^Q \cdot (\hat{F}W_i^K)^T}{\sqrt{d_k}}\right) \cdot \hat{F}W_i^V
\]
\paragraph{GGIT Integration for Global Awareness}

The Global Geographic Information Tokens (GGIT) from the AFN are integrated into the decoder via an additional cross-attention layer:
\[
q_i^{'l+1} = q_i^{l+1} + \text{CrossAttn}(q_i^{l+1}, \text{GGIT})
\]
This integration ensures that each object query has access to global scene context, enhancing detection performance particularly for objects with partial occlusion or limited visibility.

\paragraph{Mask Prediction Mechanism}

For each decoder layer $l$, we predict both a class distribution $c_i^l \in \mathbb{R}^{K+1}$ and a mask embedding $m_i^l \in \mathbb{R}^E$ for each object query $q_i^l$:
\[
\begin{aligned}
c_i^l &= \text{ClassificationHead}(q_i^l) \quad
m_i^l &= \text{MaskEmbeddingHead}(q_i^l)
\end{aligned}
\]
where $K$ is the number of object categories and the additional dimension represents a "no object" class. The mask embeddings interact with the pixel features through a dot product followed by a sigmoid activation to generate binary masks:
$
M_i^l = \sigma(m_i^l \cdot \text{PixelFeatures})
$


\subsection{Hungarian Matching and Loss Functions}

During training, we employ the Hungarian algorithm to establish a bipartite matching between predictions and ground truth instances:
$
\sigma = \arg\min_{\sigma} \sum_{i=1}^N \mathcal{L}_{\text{match}}(y_i, \hat{y}_{\sigma(i)})
$
where $\sigma$ is a permutation of $N$ elements, $y_i$ are ground truth labels and masks, and $\hat{y}_i$ are predictions. The matching cost combines classification and mask similarity:
\[
\mathcal{L}_{\text{match}}(y_i, \hat{y}_j) = \lambda_{\text{cls}} \mathcal{L}_{\text{cls}}(c_i, \hat{c}_j) + \lambda_{\text{mask}} \mathcal{L}_{\text{mask}}(M_i, \hat{M}_j) + \lambda_{\text{dice}} \mathcal{L}_{\text{dice}}(M_i, \hat{M}_j)
\]
The total training loss is calculated using the matched pairs:
\[
\mathcal{L} = \sum_{i=1}^N \left[ \lambda_{\text{cls}} \mathcal{L}_{\text{cls}}(c_i, \hat{c}_{\sigma(i)}) + \lambda_{\text{mask}} \mathcal{L}_{\text{mask}}(M_i, \hat{M}_{\sigma(i)}) + \lambda_{\text{dice}} \mathcal{L}_{\text{dice}}(M_i, \hat{M}_{\sigma(i)}) \right]
\]
where $\mathcal{L}_{\text{cls}}$ is the cross-entropy loss for classification, $\mathcal{L}_{\text{mask}}$ is the binary cross-entropy loss for mask prediction, and $\mathcal{L}_{\text{dice}}$ is the Dice loss for shape similarity.

\subsection{Inference and Post-processing}

During inference, we select object queries with confidence scores above a threshold and apply a non-maximum suppression (NMS) based on mask IoU to eliminate duplicates. The final output consists of class predictions and associated binary masks:
\[
\mathcal{O} = \{(c_i, M_i) | \max_k c_{i,k} > \tau, i \in \{1,2,\ldots,N\}\}
\]
where $\tau$ is the confidence threshold and $c_{i,k}$ is the probability of query $i$ belonging to class $k$.


\section{Experiments}
\label{sec:experiments}
We conducted experiments to evaluate the performance of our proposed Co-Win algorithm, including qualitative and quantitative evaluations on three datasets. We compared our method with the popular methods and the results are shown in below.

\subsection{Datasets}

Three datasets are employed to evaluate the proposed approach: KITTI, Waymo Open Dataset, and SemanticKITTI.
The KITTI dataset includes multimodal sensor data, with this study using LiDAR and INS data. It is split into 7,481 training and 7,518 testing samples, containing 16,142 and 16,608 vehicle instances, respectively.

The Waymo Open Dataset is a large-scale dataset with LiDAR, images, and radar data, annotated with 3D bounding boxes and instance masks.  The training set includes 798,000 frames (798 sequences), and the validation set includes 202,000 frames (202 sequences), with 1,000,000 and 250,000 vehicle instances, respectively. 

SemanticKITTI provides dense point-wise annotations for the entire LiDAR field of view, with 28 classes, including dynamic and static categories. The training set contains 19,130 scans (136,374 static vehicle instances), and the validation set includes 4,071 scans (37,280 static vehicle instances).

\subsection{Evaluation Metrics}

AP50, AP70, mAP and mIoU are used to evaluate the performance of our proposed method in SemanticKITTI. AP50 and AP70 are the average precision at 50\,\% and 70\,\% IoU thresholds, respectively. mAP is the mean average precision across all average precision, while mIoU is the mean intersection over union across all classes. 
In KITTI, we use AP70 with different categories to evaluate the performance of our method. 
In Waymo Open Dataset, we adopt the official evaluation metrics: mean average precision (mAP) and mAP weighted by heading (mAPH) for a vehicle.

\subsection{Results}

Our algorithm is evaluated on all vehicle instances that are visible in the point clouds no matter the amount of occlusion. This means that at least one point of the vehicle is present in the LiDAR scan.
In \autoref{tab:sem-kitti-results}, \autoref{tab:kitti-results-70}, and \autoref{tab:waymo-results}, the above evaluation metrics are compared with previous works.

\subsection{Ablation Study}

We conducted an ablation study to evaluate the contributions of the core ideas of Co-Win. Training was conducted on the 1/2 SemanticKITTI training dataset. Evaluation was performed on the entire validation set. The baseline model is a PointPillars encoder with a ResNet-50~\cite{He2015DeepRL} backbone.
Contributions of Main Components. \autoref{tab:afpccn} and \autoref{tab:spcn} demonstrate the impact of each component of Co-Win. We evaluated performance by adding each component one by one.

\subsection{Qualitative Results}
\begin{figure}[h]
  \centering
  \includegraphics[width=\textwidth]{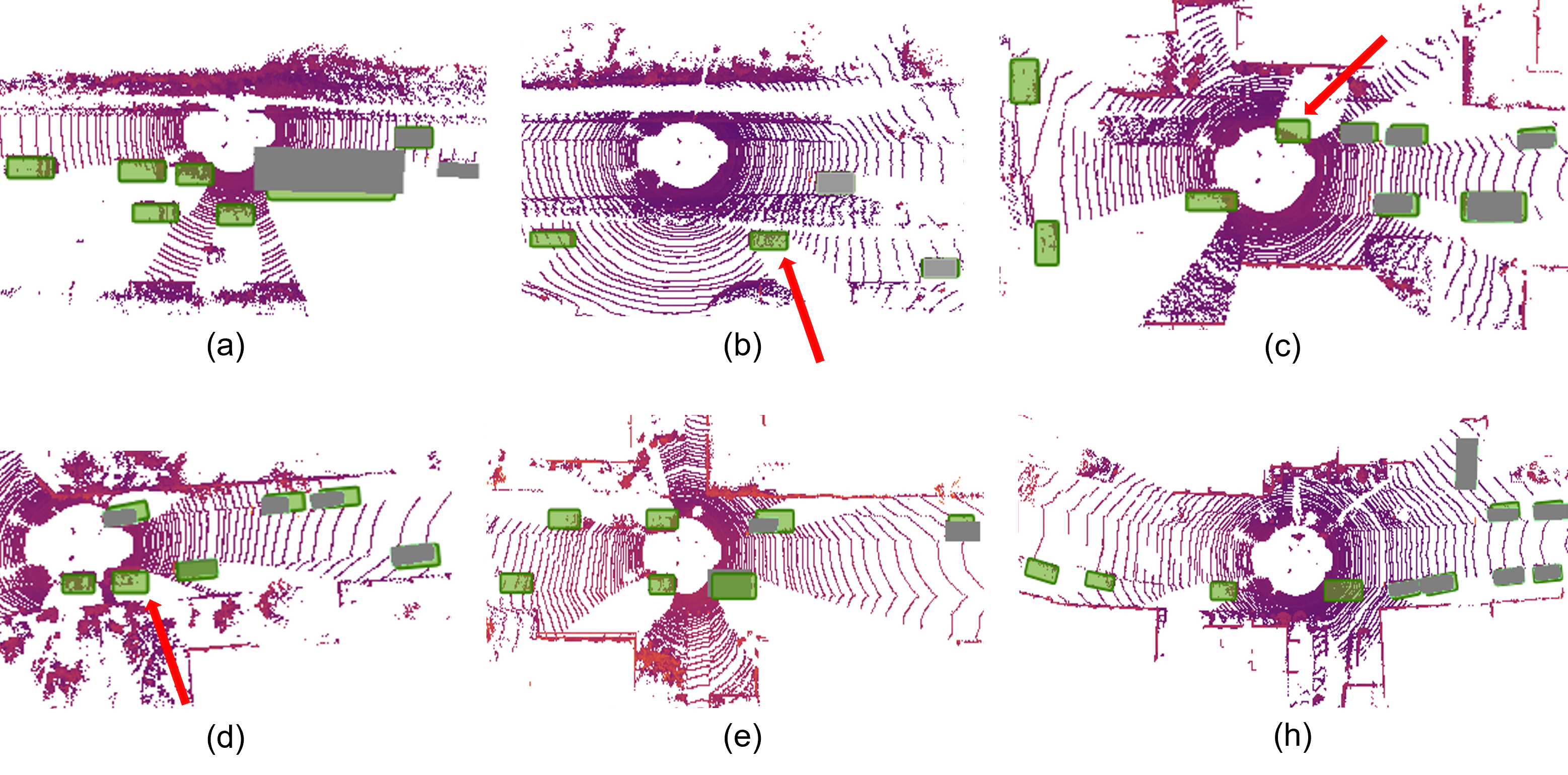}
  \caption{Qualitative test of predictions. Because the point clouds range on x-axis is [0,80], y-axis is [-40,40] and the sensor is the original center of the scenarios. Thus, the minus value on x-axis (the left part of each subfigure) will not be predicted. The green boxes are the ground truth, and the grey ones are the predictions. Comparing with the previous method~\cite{10342294}, ours shows a significant improvement (e.g. arrows in (b) (c) (d)) didn't be recognized by previous method)}
  \label{fig:finalresults}
\end{figure}

Fig.~\ref{fig:finalresults} presents the qualitative results produced by the proposed Co-win. The results are compared to the previous work~\cite{10342294}. Note that Co-win yields notably better BEV construction results than previous work. 

\section{Conclusion}
\label{sec:conclusion}

In this paper, we presented Co-Win, a novel framework for bird's-eye view perception that combines efficient point cloud processing with mask-based instance detection. 
Our work makes three significant technical contributions:
First, we introduced the Axis-Fusion Point Cloud Compact Network (AFN), a specialized encoder that transforms raw point clouds into structured BEV representations. By analyzing point clouds from three orthogonal projection planes and employing a geometric axis fusion mechanism, our encoder preserves critical spatial relationships that are typically lost in conventional voxel-based methods. Additionally, the integration of Global Geographic Information Tokens (GGIT) captures valuable scene-level context that enhances downstream processing.
Second, we developed the Sub-window Parallel Computing Network (SPCN), which implements a novel linear attention mechanism that reduces computational complexity from $O(N^2)$ to $O(N)$. This optimization enables efficient processing of large-scale BEV feature maps while maintaining representational power.
Third, we designed a mask-based decoder architecture that directly generates instance-level masks instead of axis-aligned bounding boxes. This approach captures the precise shape and orientation of objects, providing more accurate representation of irregular geometries often encountered in driving scenarios. The mask-based formulation also facilitates joint detection and segmentation, improving performance on both tasks through shared feature learning.

\clearpage

\bibliographystyle{abbrv}

\bibliography{bevref}

\end{document}